\documentclass{llncs}
\usepackage[T1]{fontenc}
\usepackage{graphicx} 
\usepackage{multirow}
\usepackage{booktabs}
\usepackage{longtable}
\usepackage[table,xcdraw]{xcolor}
\usepackage{relsize}
\usepackage{amssymb}
\usepackage{amsmath}
\usepackage{adjustbox}

\begin{document}

\setlength{\textfloatsep}{8pt} 
\setlength{\intextsep}{12pt} 


\title{AggTruth: Contextual Hallucination Detection using Aggregated Attention Scores in LLMs
\thanks{\tiny
Financed by: (1) the National Science Centre, Poland (2021/41/B/ST6/04471);
(2) CLARIN ERIC (2024–2026), funded by the Polish Minister of Science (agreement no. 2024/WK/01);
(3) CLARIN-PL, the European Regional Development Fund, FENG programme (FENG.02.04-IP.040004/24);
(4) statutory funds of the Department of Artificial Intelligence, Wroclaw Tech;
(5) the Polish Ministry of Education and Science (“International Projects Co-Funded” programme);
(6) the European Union, Horizon Europe (grant no. 101086321, OMINO);
(7) the EU project “DARIAH-PL”, under investment A2.4.1 of the National Recovery and Resilience Plan.
The views expressed are those of the authors and do not necessarily reflect those of the EU or the European Research Executive Agency.
}}

\titlerunning{AggTruth: Contextual Hallucination Detection using Aggregated...}

\author{
Piotr Matys\orcidID{0009-0004-9282-2892} \and
Jan Eliasz\orcidID{0009-0007-0851-1816}\thanks{\tiny Corresponding author: jan.eliasz@pwr.edu.pl} \and
Konrad Kiełczyński\orcidID{0009-0009-3223-2336} \and
Mikołaj Langner\orcidID{0009-0007-9531-5329} \and
Teddy Ferdinan\orcidID{0000-0003-3701-3502)} \and
Jan Kocoń\orcidID{0000-0002-7665-6896)} \and
Przemysław Kazienko\orcidID{0000-0001-5868-356X}
}
\authorrunning{Matys P., et al.}

\institute{
Department of Artificial Intelligence, Wroclaw Tech, Wrocław, Poland\\
\email{\{jan.eliasz, jan.kocon, kazienko\}@pwr.edu.pl}
}

\maketitle

\begin{abstract}
In real-world applications, Large Language Models (LLMs) often hallucinate, even in Retrieval-Augmented Generation (RAG) settings, which poses a significant challenge to their deployment. In this paper, we introduce AggTruth, a method for online detection of contextual hallucinations by analyzing the distribution of internal attention scores in the provided context (passage). Specifically, we propose four different variants of the method, each varying in the aggregation technique used to calculate attention scores. Across all LLMs examined, AggTruth demonstrated stable performance in both same-task and cross-task setups, outperforming the current SOTA in multiple scenarios. Furthermore, we conducted an in-depth analysis of feature selection techniques and examined how the number of selected attention heads impacts detection performance, demonstrating that careful selection of heads is essential to achieve optimal results.


\end{abstract}

\section{Introduction}

Large language models (LLMs) have gained popularity over the past few years due to their remarkable capabilities on a wide range of natural language processing tasks \cite{kocon2023chatgpt,peng2023rwkv,zhao2024surveylargelanguagemodels}. However, they often suffer from one major drawback, a tendency to hallucinate. Although definitions of hallucination may vary across sources, the one we adopt defines it as the phenomenon of producing responses that are non-sensical or unfaithful to the provided source content \cite{Ji_2023}. This phenomenon is a critical barrier to deploying LLMs in real-world applications where accuracy and reliability are crucial.

In our work, we specifically focus on detecting contextual hallucinations in the Retrieval-Augmented Generation (RAG) setup \cite{lewis2021retrievalaugmentedgenerationknowledgeintensivenlp}. These hallucinations can be classified as intrinsic hallucinations, which means that the model output directly conflicts with the provided context. Although RAG effectively enhances the factuality of LLMs by enriching their inputs with additional contextual information, it has its own limitations. The model may still produce hallucinations when faced with a noisy or inconsistent context or because of the inappropriate use of the accurate context \cite{Huang_2024}. Effective detection of hallucinations is also a crucial step in the self-learning LLM framework \cite{ferdinan2024unknownselflearninglargelanguage}, as it provides information on model inaccuracies. In our study, we analyzed different hallucination detection techniques and proposed some strategies. Our key contributions are as follows:
\begin{itemize}
    \item Proposing four attention map aggregation techniques allowing classifier training for windowed, real-time detection of the contextual hallucinations
    \item Conducting complex performance comparison of feature selection methods across many models and datasets
    \item Analyzing the influence of the number of heads on hallucination detection
    \item Pointing out possible improvements in the current SOTA based on the attention map method for contextual hallucination detection 
\end{itemize}

\section{Related work}

Research on hallucination detection in LLMs has evolved along several paths, with some focusing more on external response comparison methods while others examine the model's internal states during generation.

One of the simplest external methods involves assessing the factuality of a model in a zero-resource fashion by comparing inconsistencies between multiple responses \cite{manakul2023selfcheckgptzeroresourceblackboxhallucination}. A similar but more statistically grounded idea, based on semantic entropy, has also been presented \cite{Farquhar2024}. Both approaches are limited by the requirement to generate multiple answers, compromising online efficiency. 

Fine-tuning has also been proposed as a strategy to mitigate hallucinations \cite{gekhman2024doesfinetuningllmsnew}. However, while this approach is effective for specific tasks, it may lead to catastrophic forgetting \cite{ke2022continualtraininglanguagemodels} of tasks that are not included in the fine-tuning process.

 The internal state analysis approach offers an alternative perspective. For example, such analyses can include hidden states \cite{azaria-mitchell-2023-internal}. However, this article only considers a true-false dataset. Building on this foundation, \cite{chuang2024lookbacklensdetectingmitigating} showed that the attention map can also be an alternative. This approach has demonstrated better performance in transferring between different tasks while having a smaller number of input features.

\section{Contextual hallucination detection method based on attention map}

\begin{figure}[h!]
    \centering
    \includegraphics[width=0.9\textwidth]{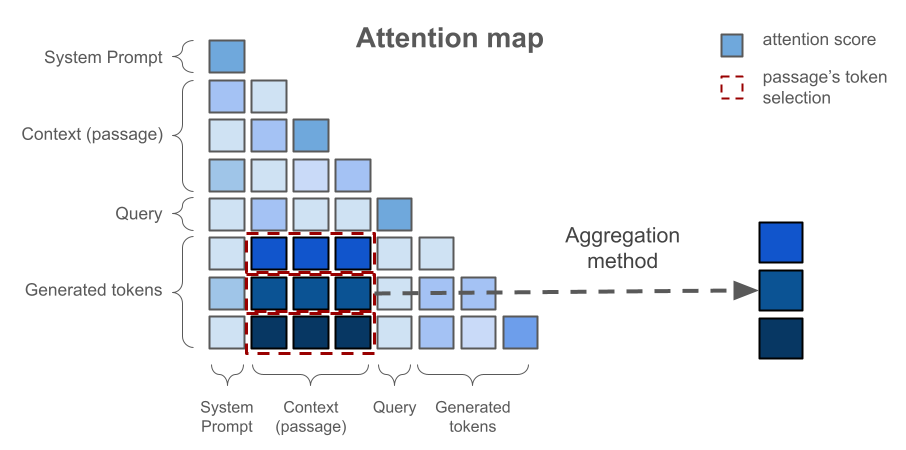}
    \caption{Selection of tokens from the LLM input
    and aggregation of attention scores. The darker 
    subregion
    indicates the attention scores of generated tokens on the provided passage. The aggregation of each such region provides one feature for the hallucination detection. Here, we have three regions resulting in three features.}
    \label{fig:attention_map}
\end{figure}

As \cite{chuang2024lookbacklensdetectingmitigating} has shown, attention map analysis provides valuable insight for LLMs hallucination detection during context-based retrieval-augmented generation (RAG) tasks. We hypothesize that examining the complete attention matrix may not be necessary. Instead, focusing specifically on the attention patterns related to the provided context (passage) could be sufficient. Such an approach should provide features that enable proper hallucination detection. We propose creating attention-based features by analyzing how newly generated tokens pay attention to the provided passage within a defined window as illustrated in Figure \ref{fig:attention_map}.

\subsection{Attention maps aggregation techniques}

For every response generated by a transformer, a four-dimensional tensor of shape $L \times H \times N \times C$ (Layers $\times$ Heads $\times$ Generated tokens $\times$ Context tokens) is created. To train a hallucination detection model, attention scores over the provided passage for each generated token have to be aggregated to reduce the map dimensionality and volume. 
For each token generated, its attention distribution across the passage is aggregated. 

Let $a_{l,h,t,i}$ be an attention score corresponding to layer $l\in\{1,\ldots,L\}$, head $h\in\{1,\ldots,H\}$, generated token $t\in\{1,\ldots,N\}$, context token $i\in\{1,\ldots,C\}$, and $\mathbf{a}_{l,h,t}$ be a vector of such scores for the whole context. The proposed aggregation techniques are as follows:

\subsubsection{AggTruth Sum}

The simplest approach to aggregate attention scores is to sum up the attention scores that were distributed over passage tokens during the generation of a given token (Equation~\ref{eq:sum_agg}).

\begin{equation}
\operatorname{Sum}_{l,h,t} =  \sum_{i=1}^{C} a_{l, h, t, i}
\label{eq:sum_agg}
\end{equation}

It is based on the assumption that LLMs produce more factual responses when they pay more attention to the provided context tokens. 
Due to the use of only tokens from the provided context, the sum of the attention scores does not equal 1, which is a desired characteristic, as it means that the examples differ from each other. 

\subsubsection{AggTruth CosSim}
In this approach, each head in a layer is described by the attention scores it assigns to the passage tokens while generating a new token. Then, the cosine similarity between the heads within the layer is calculated and the values are averaged for each head (Equation~\ref{eq:cos_sim_agg}). 

\begin{equation}
\operatorname{CosSim}_{l, h, t} = \frac{1}{H - 1} \mathlarger{\mathlarger{\sum}}_{\substack{h' = 1 \\ h' \neq h}}^{H}{ \frac{\mathbf{a}_{l,h,t}\cdot\mathbf{a}_{l,h',t}}{\lVert\mathbf{a}_{l,h,t}\rVert\lVert\mathbf{a}_{l,h',t}\rVert}}
\label{eq:cos_sim_agg}
\end{equation}

This produces a single value for each head, representing its similarity to other heads in the layer when generating a given token. The intuition behind this approach is that when unreliable content is produced, all the heads become similar to each other, indicating that the model lacks a clear focal point.

\vspace{1cm}

From the other perspective, apart from asking how much, we want to ask in which way the LLM's attention spreads across context tokens. To answer that, in our study, we treat attention scores as pseudo-probabilities. That allows us to examine the quantities related to the probability distributions. To fulfill the summing-up-to-one criteria, we do not simply normalize the attention scores. It is undesirable as it loses information about the original magnitude and distribution of attention scores. Instead, for the remaining aggregation techniques, we append an additional value to the end of the attention score vector calculated as the difference between 1 and the calculated sum of attention scores.

\subsubsection{AggTruth Entropy}
The concept of uncertainty leads to another aspect of the possible cause of hallucinations produced by LLMs. The approach described by Equation~\ref{eq:entropy_agg} measures whether a model during generation is focused only on some particular information or on the context as a whole.     

\begin{equation}
\operatorname{Entropy}_{l,h,t} = -\sum_{i=1}^C a_{l,h,t,i}\log_{2}{a_{l,h,t,i}} 
\label{eq:entropy_agg}
\end{equation}

\subsubsection{AggTruth JS-Div}
We consider hallucinations as outliers, representing a lack of knowledge required to perform the task. The main motivation behind this work was the assumption that hallucinated and non-hallucinated examples could be differentiated based on the stability of the distributions across layers, allowing for the identification of outliers. Since attention scores are treated as probability distributions, the average distribution of the heads in each layer can be calculated. The Jensen-Shannon distance is then used to measure the dissimilarity between each head and the average distribution within each layer separately (Equation~\ref{eq:js_div_agg}). 

\begin{equation}
\operatorname{JS-Div}_{l,h,t} = \sqrt{\frac{1}{2}\sum_{i=1}^{C} \left(a_{l, h, t, i}\ln{\frac{a_{l, h, t, i}}{m_{l, h, t, i}}} + a_{l, \operatorname{ref}, t, i}\ln{\frac{a_{l, \operatorname{ref}, t, i}}{m_{l, h, t, i}}}\right)}
\label{eq:js_div_agg}
\end{equation}

where:

\begin{equation*}
a_{l, \operatorname{ref}, t, i} = \frac{1}{H}\sum_{h=1}^{H} a_{l, h, t, i}, \qquad m_{l, h, t, i} = \frac{a_{l, h, t, i} + a_{l, \operatorname{ref}, t, i}}{2}
\end{equation*}

\subsection{Feature (heads) selectors}

\subsubsection{Central tendency measure ratio (Center$_\textbf{r}$)}
Any central tendency measure that can be calculated for hallucinated and not hallucinated examples can be used. The idea is to select the \textbf{r} fraction of heads with the highest (lowest) ratio between the two classes. In this study, we used the median, one of the most popular statistics for comparing groups. We keep the top \textbf{r/2} fraction of heads with the highest ratio and \textbf{r/2} fraction of heads with the lowest ratio. This technique is proposed as a naive baseline for selecting heads.

\subsubsection{Above random feature performance (Random$_\textbf{n,k}$)}
A feature with random values is appended to the dataset. Only heads with a higher absolute importance value are accepted. As we use logistic regression, we choose heads based on the values of their coefficients.
The procedure is repeated \textbf{n} times and only the heads accepted at least \textbf{k} times are selected. This is a simplified version of the Boruta feature selection algorithm \cite{Kursa2010}, which omits feature selection based on the binomial distribution and employs logistic regression instead of random forest. Using the original algorithm took much longer as more iterations were necessary, but the results were comparable.

\subsubsection{Above random feature performance - positive coefficients $\left(\text{Random}^{+}_\textbf{n,k}\right)$}
The idea is the same as above, but only positive coefficients are considered.

\subsubsection{Lasso based selection (Lasso)}
L1 regularization used by the Lasso technique shrinks some coefficients to zero, resulting in a sparse model. We leverage this property to select only the features with non-zero coefficients.

\subsubsection{Spearman correlation with the target (Spearman$_\textbf{r}$)}
During the exploratory data analysis, a high degree of multicollinearity was observed among the heads. As a result, selecting them based on the correlation among themselves or the variance inflation factor is ineffective. Thus, the selection process focuses on the correlation with the target variable and selects the \textbf{r} fraction of heads that exhibit the highest statistically significant Spearman correlation factor, with a strict threshold set at $p = 0.001$. Additionally, a setup is introduced for $r=\text{auto}$, which means that only features with a correlation greater than half of the highest observed value are considered.

\subsection{Aligning attention scores and hidden states to proper token}
The attention map returned along with a newly generated token contains information about the impact of previous tokens on its generation. However, the token under consideration does not influence the attention scores. That is, regardless of which token is selected in the generation step (which may depend on whether greedy decoding or another sampling method is used), the attention map remains the same. A token starts to influence the attention map only after it has been generated. We hypothesize that to create features that describe token \textbf{t}, the attention map extracted from the process of generating token \textbf{t+1} should be used. If this is not taken into account, each token is described by the features of its predecessor.

\subsection{Passage percentage feature}
Generating new tokens influences the number of tokens to which LLMs must pay attention. For the same input passage, the overall percentage of attention given to the passage decreases as new tokens are generated because the attention becomes more dispersed across a greater number of tokens. It may have an undesirable impact on certain attention map aggregation techniques, such as AggTruth Sum, as the aggregation results naturally decrease over time.

To address this issue, we propose incorporating an additional feature into the dataset that represents the percentage ratio of the passage length to the total input length when each token is generated. 

\section{Experimental Setup}
\begin{figure}[h!]
    \centering
    \includegraphics[width=\textwidth]{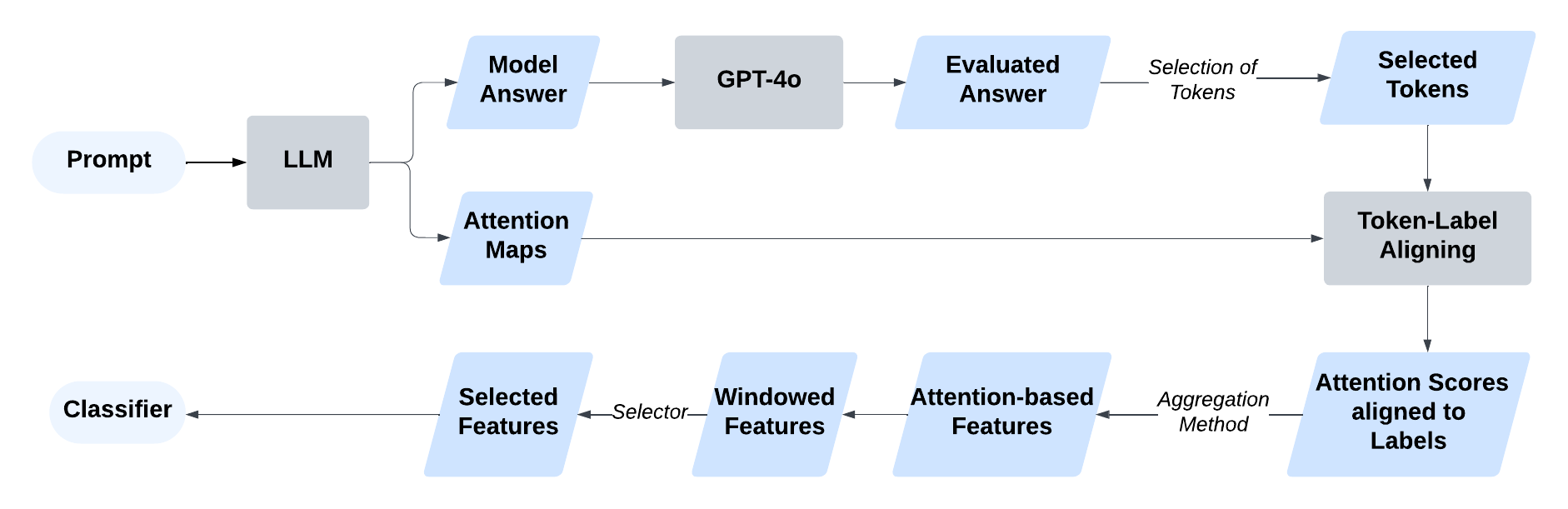}
    \caption{The whole end-to-end pipeline. It begins with a \emph{Prompt} passed to an \emph{LLM} and eventually results in \emph{Selected Features} based on which the final \emph{Classifier} detects potential hallucinations in the obtained answer.}
    \label{fig:pipeline}
\end{figure}

To produce a reliable dataset for further evaluation, we have composed a complete pipeline (Figure \ref{fig:pipeline}) that for each prompt leads to features that are eventually used to train hallucination detection classifier. The pipeline retrieves LLM answers with corresponding attention maps and then submits these answers to GPT-4o for evaluation. The process continues by selecting the attention scores put on the passage tokens and aligning each generated token with a binary label (1 for hallucinated, 0 for non-hallucinated). Attention scores are then aggregated using the chosen AggTruth aggregation technique and all generated tokens are divided into segments through windowing, i.e., they are processed into overlapping chunks of size 8, suggested by best baseline method \cite{chuang2024lookbacklensdetectingmitigating}, sliding token by token. If a chunk contains any hallucinated token, then it is aligned with label 1 and 0 otherwise. In addition, for every chunk, the features are averaged over the tokens, resulting in \textbf{a single feature representing a single head}. After processing all the examples, the results are concatenated into a unified dataset. This data then passes through a selector, producing the final feature set that enables classifiers to identify potential hallucinations within LLM responses.

\subsection{Data}

We first established a comprehensive dataset of language model responses with token-level annotations to train classifiers for hallucination detection.  We restricted our analysis to $4,096$ LLM's context length, considering that some of the models we evaluate employ the Sliding Window Attention Mechanism \cite{gemmateam2024gemma2improvingopen}, which prevents the extraction of attention scores over the whole passage once the sliding window size is exceeded. 

We focused our evaluation on two fundamental natural language processing tasks: Question Answering (QA) and Summarization. For the QA assessment, we used $1,535$ examples from Natural Questions (NQ) \cite{kwiatkowski-etal-2019-natural} and $896$ from HotPotQA \cite{yang2018hotpotqadatasetdiverseexplainable}. The summarization evaluation incorporated $1000$ examples from CNN/Daily Mail (CNN/DM)  \cite{see-etal-2017-get} and $1000$ from XSum \cite{Narayan2018DontGM}.

\begin{table}[h!]
\centering
\resizebox{\textwidth}{!}{
\begin{tabular}{lccccccc}
\toprule
\rule{0pt}{3ex} \textbf{LLM} & \textbf{Layers} & \textbf{Heads} & \textbf{ Hidden Size } & \textbf{NQ} & \textbf{HotPotQA} & \textbf{CNN/DM} & \textbf{XSum} \\ 
\midrule
\rule{0pt}{2ex}meta-llama/Llama-2-7b-chat-hf     & 32 & 32 & 4096 & 54.9 & 62.0 & 74.4 & 63.2 \\ 
meta-llama/Llama-3.1-8B-Instruct  & 32 & 32 & 4096 & 75.8 & 56.1 & 86.8 & 77.0 \\ 
unsloth/gemma-2-9b-it-bnb-4bit    & 42 & 16 & 3584 & 84.4 & 85.3 & 92.6 & 75.5 \\ 
microsoft/Phi-3.5-mini-instruct   & 32 & 32 & 3072 & 61.0 & 74.2 & 75.7 & 66.6 \\ 
\bottomrule \\
\end{tabular}
}
\caption{Configuration with key hyperparameters and truthfulness evaluation results of examined LLMs.}
\label{tab:combined_results}
\end{table}

The responses were generated using greedy decoding across the examined LLMs for all dataset examples. This process naturally produced outputs that contained both factual and hallucinated content. To establish ground truth labels at the token level, we employed GPT-4o as our judging model. We show the results of this evaluation for all models in Table \ref{tab:combined_results}. This choice was supported by recent research \cite{chuang2024lookbacklensdetectingmitigating} showing that GPT-4o's judgments align with human assessments in $97\%$ of the cases. We also conducted a study where a human expert annotated 75 samples, achieving a Cohen's Kappa of 0.7 between the expert and GPT-4o.

\begingroup

\setlength{\tabcolsep}{0.01\textwidth}

\endgroup

\subsection{Evaluation Configurations}

Our framework evaluates methods in same-task and cross-task settings. The dataset is divided into training and test sets, with the training set undergoing 5-fold cross-validation to assess model's stability and validation performance. The test set is further subdivided into subsets for the same-task and cross-task to evaluate the model's generalization. As our classifier, a Logistic Regression model with class-weight balancing is employed. This model was chosen because of its speed , interpretability and was the default model in the baselines. Time-series-specific models such as LSTM were rejected in the early testing phase due to their poor performance on the hallucination detection task. Features before being passed to the classifier are normalized using the min-max strategy.

The evaluation structure follows two primary configurations, as shown in Table \ref{tab:evaluation_datasets}. This dual-configuration approach enables us to assess the model's ability to generalize across different question-answering and summarization tasks.

\begingroup

\setlength{\tabcolsep}{0.01\textwidth}

\begin{table}[h!]
\centering
\resizebox{0.8\textwidth}{!}{
\begin{tabular}{llllll}
\toprule
\rule{0pt}{2.5ex} \multirow{2}{*}{\textbf{Source}} & \multirow{2}{*}{\textbf{Target}} & \multicolumn{2}{c}{\textbf{Source}} & \multicolumn{2}{c}{\textbf{Target}} \\
\cmidrule(lr){3-4} \cmidrule(lr){5-6}
& & \textbf{Train/Val} & \textbf{Test} & \textbf{Test(1)} & \textbf{Test(2)} \\
\cline{1-6}
\rule{0pt}{2ex} QA & SUM. & NQ & HotPotQA & XSum & CNN/DM \\
\rule{0pt}{2ex} SUM. & QA & CNN/DM & XSum & HotPotQA & NQ \\
\bottomrule \\
\end{tabular}
}
\caption{Description of datasets used at each stage of evaluation for each task setup.}
\label{tab:evaluation_datasets}
\end{table}

\endgroup
\vspace{-1.5em}

\subsection{Baselines}

\textbf{Lookback-Lens}
This method utilizes features constructed as the ratio of attention placed on the prompt versus newly generated tokens \cite{chuang2024lookbacklensdetectingmitigating}. In contrast to our approach, the attention placed on the entire provided prompt is analyzed (including the system prompt and query), which naturally may lack robustness for changing input. This method generalizes well and achieves competitive results in a cross-task setting, being the current SOTA for windowed, online hallucination detection.

\textbf{Hidden states-based classifier}
Based on the literature \cite{azaria-mitchell-2023-internal} hidden states can serve as effective features for hallucination detection. Their findings indicate that the optimal results are achieved by utilizing hidden states from the upper layers of the model architecture. Following that observation, we obtain the features of the layers $L, L-4, L-8$, where $L$ represents the total number of layers in the language model. The hidden states from these layers are processed by averaging the values across tokens within a defined window, which are then used directly as input features for the classification models.

\section{Results}

\begingroup
\setlength{\tabcolsep}{0.01\textwidth}

\begin{table}[h]
\scriptsize
\centering
\adjustbox{max width = \textwidth}{%
\begin{tabular}{p{0.2\textwidth}llcccccc}
\toprule
\rule{0pt}{2.5ex} \multirow{2}{*}{\textbf{Method}} & \multirow{2}{*}{\textbf{Source}} & \multirow{2}{*}{\textbf{Target}} & \multicolumn{3}{c}{\textbf{Source}} & \multicolumn{2}{c}{\textbf{Target}} & \multirow{2}{*}{\textbf{Gap $[\%]$}} \\
\cmidrule(lr){4-6} \cmidrule(lr){7-8}
& & & \textbf{Train} & \textbf{Val} & \textbf{Test} & \textbf{Test(1)} & \textbf{Test(2)} \\
\cline{1-9}
\multicolumn{9}{c}{\textbf{Text based NLI}} \\
\cline{1-9}
\multirow{2}{*}{SOTA NLI$^{*}$} & QA & SUM. & --- & --- & --- & --- & 0.530 & --- \\
& \cellcolor{gray!15}SUM. & \cellcolor{gray!15}QA & \cellcolor{gray!15}--- & \cellcolor{gray!15}--- & \cellcolor{gray!15}--- & \cellcolor{gray!15}--- & \cellcolor{gray!15}0.649 & \cellcolor{gray!15}--- \\
\cline{1-9}
\multicolumn{9}{c}{\textbf{Hidden States based}} \\
\cline{1-9}
\multirow{2}{*}{24th Layer} & QA & SUM. & 0.954 & 0.945 & 0.717 & 0.625 & 0.629 &  8.752 \\ 
& \cellcolor{gray!15}SUM. & \cellcolor{gray!15}QA & \cellcolor{gray!15}0.988 & \cellcolor{gray!15}0.982 & \cellcolor{gray!15}0.700 & \cellcolor{gray!15}0.707 & \cellcolor{gray!15}0.628 & \cellcolor{gray!15}5.356 \\ 
\cmidrule(r){1-3} \cmidrule(lr){4-6} \cmidrule(lr){7-8} \cmidrule(lr){9-9}
\multirow{2}{*}{28th Layer} & QA & SUM. & 0.955 & 0.946 & 0.727 & 0.603 & 0.623 & 9.564 \\ 
& \cellcolor{gray!15}SUM. & \cellcolor{gray!15}QA & \cellcolor{gray!15}0.988 & \cellcolor{gray!15}0.981 & \cellcolor{gray!15}0.691 & \cellcolor{gray!15}0.704 & \cellcolor{gray!15}0.611 & \cellcolor{gray!15}6.730 \\ 
\cmidrule(r){1-3} \cmidrule(lr){4-6} \cmidrule(lr){7-8} \cmidrule(lr){9-9}
\multirow{2}{*}{32th Layer} & QA & SUM. & 0.950 & 0.939 & 0.739 & 0.605 & 0.621 &   9.038 \\ 
& \cellcolor{gray!15}SUM. & \cellcolor{gray!15}QA & \cellcolor{gray!15}0.986 & \cellcolor{gray!15}0.978 & \cellcolor{gray!15}0.678 & \cellcolor{gray!15}0.660 & \cellcolor{gray!15}0.573 & \cellcolor{gray!15}11.117 \\ 
\cline{1-9}
\multicolumn{9}{c}{\textbf{Attention based}} \\
\cline{1-9}
\multirow{2}{0.2\textwidth}{Lookback Lens (paper)**} & QA & SUM. & --- & --- & --- & --- & 0.661 & --- \\
& \cellcolor{gray!15}SUM. & \cellcolor{gray!15}QA & \cellcolor{gray!15}--- & \cellcolor{gray!15}--- & \cellcolor{gray!15}--- & \cellcolor{gray!15}--- & \cellcolor{gray!15}0.660 & \cellcolor{gray!15}--- \\

\cmidrule(r){1-3} \cmidrule(lr){4-6} \cmidrule(lr){7-8} \cmidrule(lr){9-9}
\multirow{2}{0.2\textwidth}{Lookback Lens (classifiers)***}  
& QA & SUM. & --- & --- & 0.554 &  0.666 & 0.635  & 13.688 \\ 
& \cellcolor{gray!15}SUM. & \cellcolor{gray!15}QA & \cellcolor{gray!15}--- & \cellcolor{gray!15}--- & \cellcolor{gray!15}\textbf{0.722} & \cellcolor{gray!15}0.506 & \cellcolor{gray!15}0.506 & \cellcolor{gray!15}19.299 \\
\cmidrule(r){1-3} \cmidrule(lr){4-6} \cmidrule(lr){7-8} \cmidrule(lr){9-9}
\multirow{2}{0.2\textwidth}{Lookback Lens (retrained)****} & QA & SUM. & 0.839 & 0.833 & \textbf{0.752} & 0.643 & 0.681 & 3.952 \\ 
& \cellcolor{gray!15}SUM. & \cellcolor{gray!15}QA & \cellcolor{gray!15}0.898 & \cellcolor{gray!15}0.882 & \cellcolor{gray!15}0.700 & \cellcolor{gray!15}0.523 & \cellcolor{gray!15}0.521 & \cellcolor{gray!15}18.820 \\
\cmidrule(r){1-3} \cmidrule(lr){4-6} \cmidrule(lr){7-8} \cmidrule(lr){9-9}
\multirow{2}{0.2\textwidth}{\textit{{AggTruth Sum}}} & QA & SUM. & 0.802 & 0.799 & 0.723 & \textbf{0.670} & \textbf{0.710} &  \textbf{2.612} \\ 
& \cellcolor{gray!15}SUM. & \cellcolor{gray!15}QA & \cellcolor{gray!15}0.894 & \cellcolor{gray!15}0.885 & \cellcolor{gray!15}0.706 & \cellcolor{gray!15}\textbf{0.724} & \cellcolor{gray!15}\textbf{0.660} & \cellcolor{gray!15}\textbf{2.714} \\

\bottomrule \\
\end{tabular}
}
\caption{Comparison of Llama-2 hallucination detection efficiency (AUROC) between AggTruth Sum, SOTA NLI based method, hidden states based classifier and Lookback-Lens. \\
\\
\scriptsize{
*SOTA NLI: Results cited from \cite{chuang2024lookbacklensdetectingmitigating}. \\
**Lookback Lens (paper): Results cited from \cite{chuang2024lookbacklensdetectingmitigating}.  \\
***Lookback Lens (classifiers): Results obtained by trained classifiers shared by the authors of \cite{chuang2024lookbacklensdetectingmitigating}.    \\
****Lookback Lens (retrained): Results obtained by classifiers retrained with the script shared by the authors of \cite{chuang2024lookbacklensdetectingmitigating}, but using the same prompt as in AggTruth experimental setup. \\
}}
\label{tab:detection_results_llama2}
\end{table}

\endgroup

For each model and for each source--target pair a \textbf{Gap} value (Equation~\ref{eq:gap}) is presented apart from AUROC. 

\begin{equation}
\text{Gap}_{m} = \frac{1}{|S|} \mathlarger{\mathlarger{\sum}}_{s\in S}\frac{\max_{m'\in M}\text{AUC}_{m'}^s-\text{AUC}_m^s}{\max_{m'\in M}\text{AUC}_{m'}^s}\cdot 100\%
\label{eq:gap}
\end{equation}
where $m\in M$ is a specific method for which Gap is being calculated among all of $M$ methods that we take into account, $S$ is a set of all three test sets and $\text{AUC}_m^s$ is an AUC score obtained by method $m$ on a test set $s$. The maximum observed AUC scores across all models and test sets are presented in Table~\ref{tab:detection_best_results}. 
To~examine the stability of the methods, the Gap is averaged over the test sets. It is necessary as simply averaging AUCROC values is not informative enough, because different datasets contain varying numbers of positive samples and have different levels of difficulty, leading to different ranges of possible metric values.

Table \ref{tab:detection_results_llama2} shows the efficiency of hallucination detection for Llama-2. Unlike hidden state-based approaches, attention-based methods are less prone to overfitting to the training dataset, resulting in a higher AUCROC on out-of-domain target datasets. However, for Lookback-Lens, a substantial decrease in performance can be observed for SUM. $\to$ QA setting for both the provided by authors and retrained classifiers compared to the results of \cite{chuang2024lookbacklensdetectingmitigating}. 
AggTruth Sum achieves the highest AUCROC across all target tasks, while also getting stable, high AUCROC on the source test setting, resulting in the lowest Gap values. This observation suggests that AggTruth can be more robust to the choice of the training data.

Table \ref{tab:detection_results_other_models} presents the results for other models. It lacks the Lookback-Lens method, as the code provided by its authors is not easily adaptable for models other than Llama-2. Considering the gap values, AggTruth JS-DIV performed best for Phi-3.5 and Llama-3, while AggTruth Sum led on Gemma-2. Overall, AggTruth achieves the highest AUCROC and the lowest gap values. It should be noted that, even though the hidden state-based approach performed best in QA $\to$ SUM. task for Gemma-2, on SUM. $\to$ QA it performs worse than a random classifier, which is proof of the instability of this approach.

\begin{table}[h]
\scriptsize
\adjustbox{max width = \textwidth}{%
\begin{tabular}{lp{0.2\textwidth}llcccccc}
\toprule
\rule{0pt}{2.5ex} \multirow{2}{*}{\textbf{LLM}} & \multirow{2}{*}{\textbf{Method}} & \multirow{2}{*}{\textbf{Source}} & \multirow{2}{*}{\textbf{Target}} & \multicolumn{3}{c}{\textbf{Source}} & \multicolumn{2}{c}{\textbf{Target}} & \multirow{2}{*}{\textbf{Gap $[\%]$}} \\
\cmidrule(lr){5-7} \cmidrule(lr){8-9}
& & & & \textbf{Train} & \textbf{Val} & \textbf{Test} & \textbf{Test(1)} & \textbf{Test(2)} \\
\midrule
\rule{0pt}{3ex} \multirow{8}{0.15\textwidth}{llama-3.1-8B- Instruct} 
& \multirow{2}{0.2\textwidth}{Hidden States (24th Layer)} & QA & SUM. & 0.932 & 0.925 & 0.877 & 0.606 & 0.623 & 8.957 \\
& & \cellcolor{gray!15}SUM. & \cellcolor{gray!15}QA & \cellcolor{gray!15}0.996 & \cellcolor{gray!15}0.991 & \cellcolor{gray!15}0.680 & \cellcolor{gray!15}\textbf{0.860} & \cellcolor{gray!15}0.768 & \cellcolor{gray!15}5.049 \\
\cmidrule(r){2-4} \cmidrule(lr){5-7} \cmidrule(lr){8-9} \cmidrule(lr){10-10}
& \multirow{2}{0.2\textwidth}{Hidden States (28th Layer)} & QA & SUM. & 0.932 & 0.925 & 0.885 & 0.607 & 0.619 & 8.745 \\
& & \cellcolor{gray!15}SUM. & \cellcolor{gray!15}QA & \cellcolor{gray!15}0.997 & \cellcolor{gray!15}0.993 & \cellcolor{gray!15}0.665 & \cellcolor{gray!15}0.851 & \cellcolor{gray!15}0.761 & \cellcolor{gray!15}6.439 \\
\cmidrule(r){2-4} \cmidrule(lr){5-7} \cmidrule(lr){8-9} \cmidrule(lr){10-10}
& \multirow{2}{0.2\textwidth}{Hidden States (32th Layer)} & QA & SUM. & 0.927 & 0.921 & \textbf{0.886} & 0.582 & 0.589 & 11.314 \\
& & \cellcolor{gray!15}SUM. & \cellcolor{gray!15}QA & \cellcolor{gray!15}0.996 & \cellcolor{gray!15}0.990 & \cellcolor{gray!15}0.633 & \cellcolor{gray!15}0.844 & \cellcolor{gray!15}0.751 & \cellcolor{gray!15}8.573 \\
\cline{2-10}
& \multirow{2}{0.2\textwidth}{\textit{{AggTruth JS-Div}}} & QA & SUM. & 0.854 & 0.851 & 0.845 & \textbf{0.689} & \textbf{0.722} & \textbf{1.554} \\
& & \cellcolor{gray!15}SUM. & \cellcolor{gray!15}QA & \cellcolor{gray!15}0.893 & \cellcolor{gray!15}0.875 & \cellcolor{gray!15}\textbf{0.715} & \cellcolor{gray!15}0.766 & \cellcolor{gray!15}\textbf{0.853} & \cellcolor{gray!15}\textbf{3.786} \\
\midrule
\rule{0pt}{3ex} \multirow{8}{0.15\textwidth}{phi-3.5-mini- instruct} 
& \multirow{2}{0.2\textwidth}{Hidden States (24th Layer)} & QA & SUM. & 0.939 & 0.927 & 0.702 & 0.611 & 0.634 & 6.401 \\
& & \cellcolor{gray!15}SUM. & \cellcolor{gray!15}QA & \cellcolor{gray!15}0.967 & \cellcolor{gray!15}0.953 & \cellcolor{gray!15}0.670 & \cellcolor{gray!15}0.523 & \cellcolor{gray!15}0.648 & \cellcolor{gray!15}11.161 \\
\cmidrule(r){2-4} \cmidrule(lr){5-7} \cmidrule(lr){8-9} \cmidrule(lr){10-10}
& \multirow{2}{0.2\textwidth}{Hidden States (28th Layer)} & QA & SUM. & 0.939 & 0.928 & 0.707 & 0.601 & 0.632 & 6.753 \\
& & \cellcolor{gray!15}SUM. & \cellcolor{gray!15}QA & \cellcolor{gray!15}0.968 & \cellcolor{gray!15}0.953 & \cellcolor{gray!15}0.662 & \cellcolor{gray!15}0.559 & \cellcolor{gray!15}\textbf{0.649} & \cellcolor{gray!15}9.867 \\
\cmidrule(r){2-4} \cmidrule(lr){5-7} \cmidrule(lr){8-9} \cmidrule(lr){10-10}
& \multirow{2}{0.2\textwidth}{Hidden States (32th Layer)} & QA & SUM. & 0.933 & 0.920 & \textbf{0.720} & 0.593 & 0.620 & 7.215 \\
& & \cellcolor{gray!15}SUM. & \cellcolor{gray!15}QA & \cellcolor{gray!15}0.967 & \cellcolor{gray!15}0.952 & \cellcolor{gray!15}0.648 & \cellcolor{gray!15}0.555 & \cellcolor{gray!15}0.614 & \cellcolor{gray!15}12.503 \\
\cline{2-10}
& \multirow{2}{0.2\textwidth}{\textit{{AggTruth JS-Div}}} & QA & SUM. & 0.797 & 0.790 & 0.712 & \textbf{0.674} & \textbf{0.647} & \textbf{2.172} \\
& & \cellcolor{gray!15}SUM. & \cellcolor{gray!15}QA & \cellcolor{gray!15}0.849 & \cellcolor{gray!15}0.835 & \cellcolor{gray!15}\textbf{0.703} & \cellcolor{gray!15}\textbf{0.702} & \cellcolor{gray!15}0.636 & \cellcolor{gray!15}\textbf{2.029} \\
\midrule
\rule{0pt}{3ex} \multirow{8}{0.15\textwidth}{gemma-2-9b- it-bnb-4bit} 
& \multirow{2}{0.2\textwidth}{Hidden States (34th Layer)} & QA & SUM. & 0.961 & 0.952 & 0.812 & 0.579 & \textbf{0.644} & \textbf{4.383} \\
& & \cellcolor{gray!15}SUM. & \cellcolor{gray!15}QA & \cellcolor{gray!15}0.999 & \cellcolor{gray!15}0.996 & \cellcolor{gray!15}0.582 & \cellcolor{gray!15}0.604 & \cellcolor{gray!15}0.610 & \cellcolor{gray!15}14.497 \\
\cmidrule(r){2-4} \cmidrule(lr){5-7} \cmidrule(lr){8-9} \cmidrule(lr){10-10}
& \multirow{2}{0.2\textwidth}{Hidden States (38th Layer)} & QA & SUM. & 0.961 & 0.952 & 0.828 & 0.569 & 0.630 & 4.946 \\
& & \cellcolor{gray!15}SUM. & \cellcolor{gray!15}QA & \cellcolor{gray!15}0.999 & \cellcolor{gray!15}0.996 & \cellcolor{gray!15}0.586 & \cellcolor{gray!15}0.443 & \cellcolor{gray!15}0.537 & \cellcolor{gray!15}24.891 \\
\cmidrule(r){2-4} \cmidrule(lr){5-7} \cmidrule(lr){8-9} \cmidrule(lr){10-10}
& \multirow{2}{0.2\textwidth}{Hidden States (42th Layer)} & QA & SUM. & 0.958 & 0.948 & \textbf{0.824} & 0.563 & 0.610 & 6.426 \\
& & \cellcolor{gray!15}SUM. & \cellcolor{gray!15}QA & \cellcolor{gray!15}0.998 & \cellcolor{gray!15}0.995 & \cellcolor{gray!15}0.582 & \cellcolor{gray!15}0.316 & \cellcolor{gray!15}0.482 & \cellcolor{gray!15}33.292 \\
\cline{2-10}
& \multirow{2}{0.2\textwidth}{\textit{{AggTruth Sum}}} & QA & SUM. & 0.835 & 0.827 & 0.742 & \textbf{0.625} & 0.609 & 6.606 \\
& & \cellcolor{gray!15}SUM. & \cellcolor{gray!15}QA  & \cellcolor{gray!15}0.745 & \cellcolor{gray!15}0.736	& \cellcolor{gray!15}\textbf{0.617}	& \cellcolor{gray!15}\textbf{0.749}	& \cellcolor{gray!15}\textbf{0.625}	& \cellcolor{gray!15}\textbf{5.901} \\
\bottomrule \\
\end{tabular}
}
\caption{Results of hallucination detection for other LLMs that compare our approach with hidden states (baseline).}
\label{tab:detection_results_other_models}
\end{table}

The best results for each AggTruth aggregation method grouped by task are presented in appendix Table~\ref{tab:detection_results_all}.

\section{Predictive Power of Different Heads}
In \cite{chuang2024lookbacklensdetectingmitigating}, the authors showed that increasing the number of heads passed to the classifier increases the performance, and taking all of them produces the best results. However, they focused on a very small subset of heads (10, 50, and 100) for Llama-2 which is approximately $1\%$, $5\%$, and $10\%$ of all its heads. Additionally, they checked it only on responses divided into predefined spans (in contrast to windows) that are not available during the decoding.
 One of the most important questions we wanted to answer was whether we would observe a similar percentage of heads needed to achieve the best performance for each of the LLMs and what the optimal number of them is. In Table \ref{tab:heads_selection} we show the results for the detection tasks, comparing the use of all heads versus a percentage of them across different LLMs. Here, feature selection can differ between tasks, as it is a training dataset-specific setting and as different training data can contain various amounts of predictive information. We state that not taking all heads can not only reduce the training time and complexity of the model but also improve the performance. The best performance was observed when taking at least around half of the heads for all LLMs, but it is possible to use even $10-20\%$ of them for some tasks without a significant decrease. In some experiments, that low number of heads was sufficient to obtain higher AUCROC than when using all of them. As we did not perform hypertuning and only ran specific configurations, we suppose that for specific datasets even better results can be achieved.

 \begingroup

\begin{table}[h]
\scriptsize
\adjustbox{max width = \textwidth}{%
\begin{tabular}{p{0.15\textwidth}llllccccc}
\toprule
\rule{0pt}{2.5ex} \textbf{LLM} & \textbf{AggTruth} & \textbf{Source} & \textbf{Target} & \textbf{Selector} & \textbf{Heads [\%]} & \textbf{Test} & \textbf{Test(1)} & \textbf{Test(2)} & \textbf{Gap [\%]} \\
\midrule
\multirow{6}{0.15\textwidth}{llama-2-7b-chat-hf} & \multirow{6}{*}{Sum} & \multirow{3}{*}{QA} & \multirow{3}{*}{SUM.} & --- & 100.0 & 0.727 & 0.582 & 0.697 & 6.836 \\
& & & & Random$^{+}_{3,3}$ & 48.4 & 0.723 & 0.670 & 0.710 & \textbf{2.048} \\
& & & & Center$_{0.2}$ & \textbf{19.9} & 0.713 & 0.638 & 0.739 & 2.798 \\
\cline{3-10}
& &\cellcolor{gray!15}&\cellcolor{gray!15}&\cellcolor{gray!15}--- & \cellcolor{gray!15}100.0 & \cellcolor{gray!15}0.705 & \cellcolor{gray!15}0.722 & \cellcolor{gray!15}0.660 & \cellcolor{gray!15}2.159 \\
& & \cellcolor{gray!15}SUM.& \cellcolor{gray!15}QA& \cellcolor{gray!15}Spearman$_{1.0}$& \cellcolor{gray!15}96.2 & \cellcolor{gray!15}0.706 & \cellcolor{gray!15}0.724 & \cellcolor{gray!15}0.660 & \cellcolor{gray!15}\textbf{1.975} \\
& &\cellcolor{gray!15} &\cellcolor{gray!15} & \cellcolor{gray!15}Random$^{+}_{3,3}$ & \cellcolor{gray!15}\textbf{45.6} & \cellcolor{gray!15}0.699 & \cellcolor{gray!15}0.692 & \cellcolor{gray!15}0.639 & \cellcolor{gray!15}4.776 \\
\cline{1-10}
\multirow{6}{0.15\textwidth}{llama-3.1-8B-Instruct} & \multirow{6}{*}{JS-Div} & \multirow{3}{*}{QA} & \multirow{3}{*}{SUM.} & --- & 100.0 & 0.844 & 0.687 & 0.720 & 1.740 \\
& & & & Random$_{3,2}$ & 96.8 & 0.845 & 0.689 & 0.722 & \textbf{1.554} \\
& & & & Lasso & \textbf{19.2} & 0.856 & 0.685 & 0.708 & 1.973 \\
\cline{3-10}
& & \cellcolor{gray!15} & \cellcolor{gray!15} & \cellcolor{gray!15}--- & \cellcolor{gray!15}100.0 & \cellcolor{gray!15}0.714 & \cellcolor{gray!15}0.765 & \cellcolor{gray!15}0.854 & \cellcolor{gray!15}3.847 \\
& &\cellcolor{gray!15}SUM. &\cellcolor{gray!15}QA  & \cellcolor{gray!15}Random$_{3,1}$ & \cellcolor{gray!15}86.7 & \cellcolor{gray!15}0.715 & \cellcolor{gray!15}0.766 & \cellcolor{gray!15}0.853 & \cellcolor{gray!15}\textbf{3.786} \\
& &\cellcolor{gray!15} &\cellcolor{gray!15} & \cellcolor{gray!15}Random$^{+}_{3,3}$ & \cellcolor{gray!15}\textbf{43.2} & \cellcolor{gray!15}0.702 & \cellcolor{gray!15}0.770 & \cellcolor{gray!15}0.847 & \cellcolor{gray!15}4.512 \\
\cline{1-10}
\multirow{6}{0.15\textwidth}{phi-3.5-mini-instruct} & \multirow{6}{*}{JS-Div} & \multirow{3}{*}{QA} & \multirow{3}{*}{SUM.} & --- & 100.0 & 0.715 & 0.651 & 0.633 & 3.905 \\
& & & & Spearman$_{1.0}$& 76.6 & 0.712 & 0.674 & 0.647 & \textbf{2.172} \\
& & & & Center$_{0.2}$ & \textbf{19.9} & 0.659 & 0.660 & 0.663 & 4.406 \\
\cline{3-10}
& & \cellcolor{gray!15} &\cellcolor{gray!15} & \cellcolor{gray!15}--- & \cellcolor{gray!15}100.0 & \cellcolor{gray!15}0.703 & \cellcolor{gray!15}0.702 & \cellcolor{gray!15}0.636 & \cellcolor{gray!15}\textbf{2.029} \\
& &\cellcolor{gray!15}SUM. &\cellcolor{gray!15}QA & \cellcolor{gray!15}Random$_{3,1}$ & \cellcolor{gray!15}95.2 & \cellcolor{gray!15}0.702 & \cellcolor{gray!15}0.703 & \cellcolor{gray!15}0.635 & \cellcolor{gray!15}2.038 \\
& &\cellcolor{gray!15} &\cellcolor{gray!15} & \cellcolor{gray!15}Spearman$_{\text{auto}}$ & \cellcolor{gray!15}\textbf{31.3} & \cellcolor{gray!15}0.696 & \cellcolor{gray!15}0.713 & \cellcolor{gray!15}0.633 & \cellcolor{gray!15}2.045 \\
\cline{1-10}
\multirow{6}{0.15\textwidth}{gemma-2-9b-it-bnb-4bit} & \multirow{6}{*}{Sum} & \multirow{3}{*}{QA} & \multirow{3}{*}{SUM.} & --- & 100.0 & 0.729 & 0.637 & 0.590 & 7.512 \\
& & & & Center$_{0.5}$ & 49.9 & 0.742 & 0.625 & 0.609 & \textbf{6.606} \\
& & & & Spearman$_{0.2}$ & \textbf{19.9} & 0.759 & 0.583 & 0.634 & 6.819 \\
\cline{3-10}
& & \cellcolor{gray!15} & \cellcolor{gray!15} & \cellcolor{gray!15}--- & \cellcolor{gray!15}100.0 & \cellcolor{gray!15}0.662 & \cellcolor{gray!15}0.697 & \cellcolor{gray!15}0.572 & \cellcolor{gray!15}8.642 \\
& &\cellcolor{gray!15}SUM. &\cellcolor{gray!15}QA & \cellcolor{gray!15}Spearman$_{0.5}$ & \cellcolor{gray!15}49.9 & \cellcolor{gray!15}0.659 & \cellcolor{gray!15}0.748 & \cellcolor{gray!15}0.593 & \cellcolor{gray!15}\textbf{5.563} \\
& &\cellcolor{gray!15} &\cellcolor{gray!15} & \cellcolor{gray!15}Spearman$_{0.1}$ & \cellcolor{gray!15}\textbf{10.0} & \cellcolor{gray!15}0.617 & \cellcolor{gray!15}0.749 & \cellcolor{gray!15}0.625 & \cellcolor{gray!15}5.901 \\
\bottomrule \\
\end{tabular}
}
\caption{Influence of the percentage of selected heads on results with different selectors. Comparisons are made for each LLM, aggregation method, and task between no feature selection (marked with hyphens), best-obtained selection method w.r.t. Gap value, and method resulted in the lowest number of heads.}
\label{tab:heads_selection}
\end{table}

\endgroup

During the study, we were interested in examining the predictive power of the proposed features to differentiate samples and investigate why selecting features with positive coefficients can be so effective. For each examined dataset and LLM, we calculated the mean value of each feature separately for a group of the non-hallucinated and the hallucinated.
\begin{figure}[h!]
    \centering
    \includegraphics[width=0.9\linewidth]{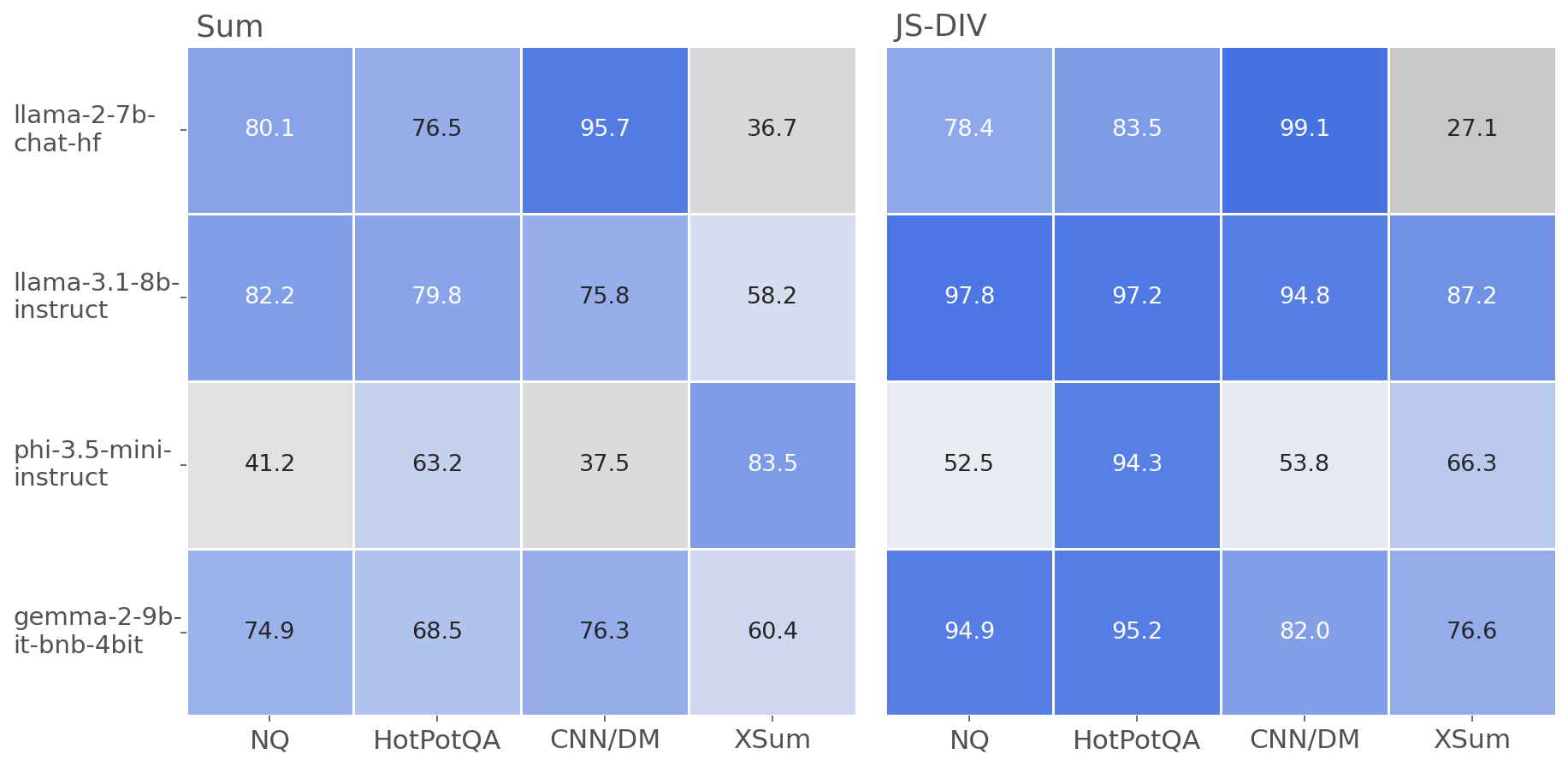}
    \caption{Percentage of heads for which the mean features (i.e. AggTruth Sum/JS-Div) are statistically higher for non-hallucinated examples than for the hallucinated ones. The comparison was conducted using Welch’s t--test with p--value = 0.01} 
    \label{fig:welch_test_results}
\end{figure}

For each head, we conducted Welch's t--test (with p--value$=0.01$) to check whether a head evinces higher mean feature values for non-hallucinated examples than for hallucinated ones. Figure \ref{fig:welch_test_results} shows the percentages of heads that meet this condition. We observed that for most used datasets and LLMs both mean Sum and mean Jensen-Shannon distance for the non-hallucinated examples were statistically higher than for the hallucinated ones. For some datasets, almost all means for non-hallucinated examples were higher. 
Due to a relatively small number of datasets and observations, we leave it for further investigation. Nevertheless, we would like to point out some characteristics that can potentially differentiate these hallucinated samples in an effective way.

\begin{figure}[h!]
    \centering
    \includegraphics[width=\linewidth]{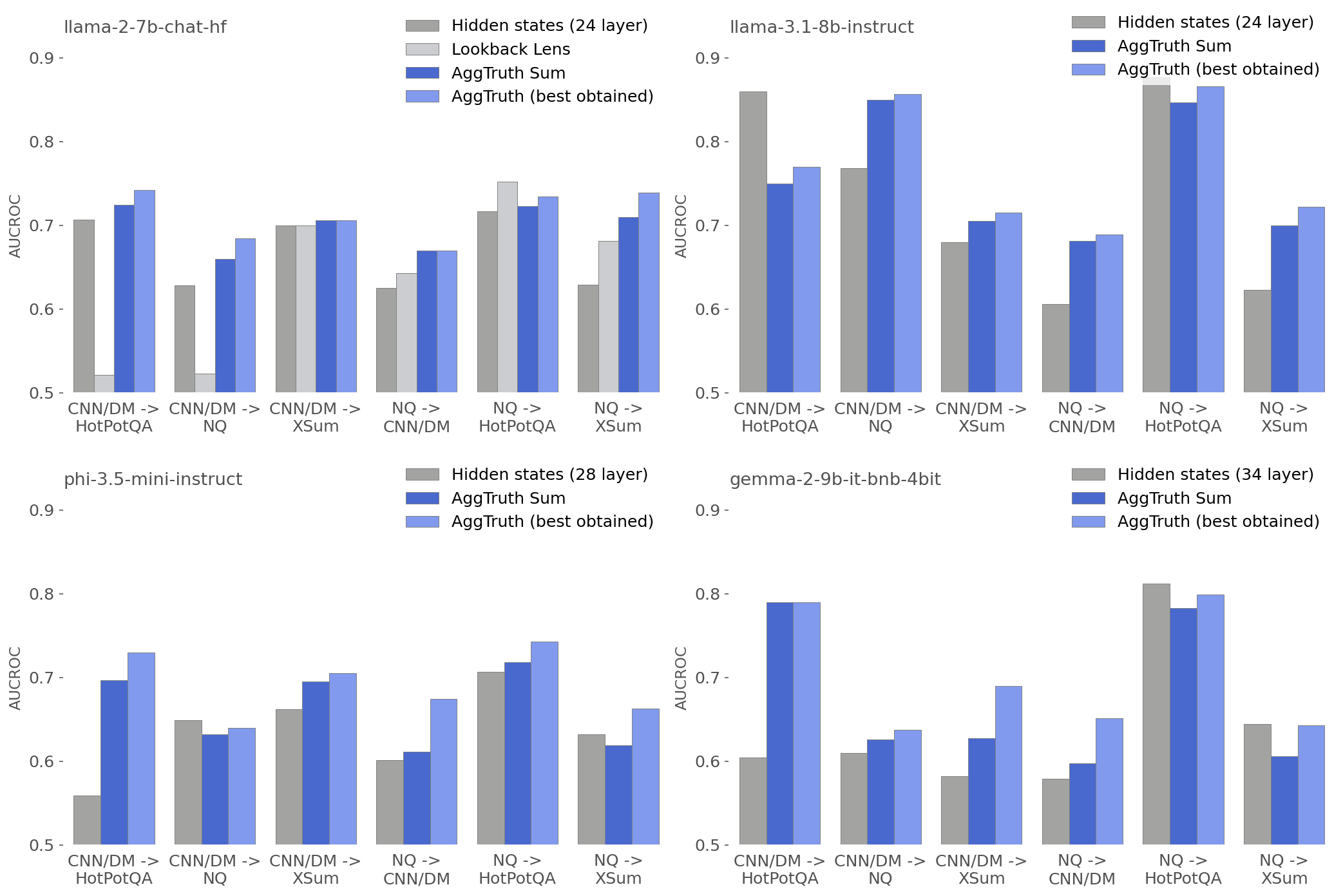}
    \caption{AUCROC hallucination detection results grouped by LLMs and tasks for best obtained hidden states-based method w.r.t. Gap value, Lookback Lens (only for Llama-2), AggTruth Sum and best obtained AggTruth method for a specific dataset and LLM.}
    \label{fig:sum_baselines_comparison}
\end{figure}

 When selecting an aggregation method for LLM, careful consideration is needed since different models may perform optimally with different methods. Based on our research findings, we recommend the AggTruth Sum as the preferred default method. It is the most stable and robust approach that resulted in the lowest overall gap across all LLMs examined and tasks, outperforming baselines on the majority of datasets, which is shown in Figure \ref{fig:sum_baselines_comparison}. More importantly, calculating the sum of the attention scores is the fastest method across all the proposed aggregations. This performance benefit is particularly crucial in classifier-guided decoding scenarios, where the most truthful generation path is selected \cite{chuang2024lookbacklensdetectingmitigating}. It is also the most intuitive and explainable approach which should be preferred when performance is satisfactory. The feature selection method is more defined by the training set, but our experiments show that the Spearman selector can be used as a first guess. Alternatively, the above random feature selector (potentially taking only positive coefficients) can be applied.
\section {Conclusions}
We proposed a novel approach to effective online contextual hallucination detection, called AggTruth, with four possible aggregation methods based on attention scores of context tokens. Our method outperforms other baselines, achieving high and stable results across all examined datasets and various LLMs. Moreover, the presented feature selection methods not only reduce the number of heads needed for classification but also improve overall detection performance. 

We believe that analyzing specific parts of attention scores vectors can open up new opportunities to detect hallucinations not only in RAG settings but also for other tasks. By design, the AggTruth detector is indicated to be used during the LLM decoding step. 
AggTruth also has its natural limitations, such as the access to context tokens' attention scores for generated tokens, which are not available when using flash-attention. Additionally, some attention patterns, such as sparse or window attention, can limit the use of AggTruth. However, such optimizations, in addition to examining a window of size one and different aggregation methods, are intended for future work along with the development of a novel dedicated classifier-guided decoding method.


\bibliographystyle{splncs04}
\bibliography{main}

\appendix

\section{Appendix}

\begin{table}[h]
\centering
\scriptsize
\adjustbox{max width = \textwidth}{%
\begin{tabular}{p{0.15\textwidth}llllcccccc}
\toprule
\rule{0pt}{2.5ex} \multirow{2}{*}{\textbf{LLM}} & \multirow{2}{*}{\textbf{AggTruth}} & \multirow{2}{*}{\textbf{Selector}} & \multirow{2}{*}{\textbf{Source}} & \multirow{2}{*}{\textbf{Target}} & \multicolumn{3}{c}{\textbf{Source}} & \multicolumn{2}{c}{\textbf{Target}} & \multirow{2}{*}{\textbf{Gap [\%]}} \\
\cmidrule(lr){6-8} \cmidrule(lr){9-10}
& & & & & \textbf{Train} & \textbf{Val} & \textbf{Test} & \textbf{Test (1)} & \textbf{Test (2)} \\
\midrule
\multirow{8}{0.15\textwidth}{llama-2-7b-chat-hf} & \multirow{2}{*}{CosSim} & Lasso & QA & SUM. & 0.858 & 0.853 & \textbf{0.731} & 0.587 & 0.713 & 5.639 \\
& & \cellcolor{gray!15}Spearman$_{1.0}$& \cellcolor{gray!15}SUM. & \cellcolor{gray!15}QA & \cellcolor{gray!15}0.884 & \cellcolor{gray!15}0.875 & \cellcolor{gray!15}0.659 & \cellcolor{gray!15}0.716 & \cellcolor{gray!15}\textbf{0.684} & \cellcolor{gray!15}3.380 \\
\cmidrule(r){2-5} \cmidrule(lr){6-8} \cmidrule(lr){9-11}
& \multirow{2}{*}{Entropy} & Random$^{+}_{3,3}$ & QA & SUM. & 0.807 & 0.805 & 0.726 & 0.651 & \textbf{0.721} & 2.393 \\
& & \cellcolor{gray!15}Spearman$_{1.0}$& \cellcolor{gray!15}SUM. & \cellcolor{gray!15}QA & \cellcolor{gray!15}0.894 & \cellcolor{gray!15}0.885 & \cellcolor{gray!15}0.699 & \cellcolor{gray!15}\textbf{0.742} & \cellcolor{gray!15}0.674 & \cellcolor{gray!15}\textbf{0.844} \\
\cmidrule(r){2-5} \cmidrule(lr){6-8} \cmidrule(lr){9-11}
& \multirow{2}{*}{JS-Div} & Center$_{0.2}$ & QA & SUM. & 0.767 & 0.765 & 0.727 & 0.639 & \textbf{0.721} & 2.911 \\
& & \cellcolor{gray!15}Spearman$_{\text{auto}}$ & \cellcolor{gray!15}SUM. & \cellcolor{gray!15}QA & \cellcolor{gray!15}0.839 & \cellcolor{gray!15}0.833 & \cellcolor{gray!15}0.674 & \cellcolor{gray!15}0.684 & \cellcolor{gray!15}0.609 & \cellcolor{gray!15}7.793 \\
\cmidrule(r){2-5} \cmidrule(lr){6-8} \cmidrule(lr){9-11}
& \multirow{2}{*}{Sum} & Random$^{+}_{3,3}$ & QA & SUM. & 0.802 & 0.799 & 0.723 & \textbf{0.670} & 0.710 & \textbf{2.048} \\
& & \cellcolor{gray!15}Spearman$_{1.0}$& \cellcolor{gray!15}SUM. & \cellcolor{gray!15}QA & \cellcolor{gray!15}0.894 & \cellcolor{gray!15}0.885 & \cellcolor{gray!15}\textbf{0.706} & \cellcolor{gray!15}0.724 & \cellcolor{gray!15}0.660 & \cellcolor{gray!15}1.975 \\
\midrule
\multirow{8}{0.15\textwidth}{llama-3.1-8B-Instruct} & \multirow{2}{*}{CosSim} & Cente$r_{0.1}$ & QA & SUM. & 0.799 & 0.797 & 0.849 & 0.641 & 0.687 & 5.336 \\
& & \cellcolor{gray!15}Random$_{3,3}$ & \cellcolor{gray!15}SUM. & \cellcolor{gray!15}QA & \cellcolor{gray!15}0.948 & \cellcolor{gray!15}0.930 & \cellcolor{gray!15}0.667 & \cellcolor{gray!15}0.755 & \cellcolor{gray!15}0.846 & \cellcolor{gray!15}6.747 \\
\cmidrule(r){2-5} \cmidrule(lr){6-8} \cmidrule(lr){9-11}
& \multirow{2}{*}{Entropy} & Spearman$_{\text{auto}}$ & QA & SUM. & 0.862 & 0.858 & \textbf{0.861} & 0.638 & 0.632 & 7.551 \\
& & \cellcolor{gray!15}Spearman$_{1.0}$& \cellcolor{gray!15}SUM. & \cellcolor{gray!15}QA & \cellcolor{gray!15}0.896 & \cellcolor{gray!15}0.878 & \cellcolor{gray!15}0.691 & \cellcolor{gray!15}0.750 & \cellcolor{gray!15}0.851 & \cellcolor{gray!15}5.622 \\
\cmidrule(r){2-5} \cmidrule(lr){6-8} \cmidrule(lr){9-11}
& \multirow{2}{*}{JS-Div} & Random$_{3,2}$ & QA & SUM. & 0.854 & 0.851 & 0.845 & \textbf{0.689} & \textbf{0.722} & \textbf{1.554} \\
& & \cellcolor{gray!15}Random$_{3,3}$ & \cellcolor{gray!15}SUM. & \cellcolor{gray!15}QA & \cellcolor{gray!15}0.893 & \cellcolor{gray!15}0.875 & \cellcolor{gray!15}\textbf{0.715} & \cellcolor{gray!15}\textbf{0.766} & \cellcolor{gray!15}\textbf{0.853} & \cellcolor{gray!15}\textbf{3.786} \\
\cmidrule(r){2-5} \cmidrule(lr){6-8} \cmidrule(lr){9-11}
& \multirow{2}{*}{Sum} & Spearman$_{1.0}$& QA & SUM. & 0.870 & 0.866 & 0.847 & 0.681 & 0.700 & 2.883 \\
& & \cellcolor{gray!15}--- & \cellcolor{gray!15}SUM. & \cellcolor{gray!15}QA & \cellcolor{gray!15}0.928 & \cellcolor{gray!15}0.910 & \cellcolor{gray!15}0.705 & \cellcolor{gray!15}0.750 & \cellcolor{gray!15}0.850 & \cellcolor{gray!15}4.979 \\
\midrule
\multirow{8}{0.15\textwidth}{phi-3.5-mini-instruct} & \multirow{2}{*}{CosSim} & Spearman$_{\text{auto}}$ & QA & SUM. & 0.666 & 0.663 & 0.671 & 0.575 & 0.578 & 12.421 \\
& & \cellcolor{gray!15}Lasso & \cellcolor{gray!15}SUM. & \cellcolor{gray!15}QA & \cellcolor{gray!15}0.812 & \cellcolor{gray!15}0.799 & \cellcolor{gray!15}0.647 & \cellcolor{gray!15}\textbf{0.709} & \cellcolor{gray!15}0.623 & \cellcolor{gray!15}5.069 \\
\cmidrule(r){2-5} \cmidrule(lr){6-8} \cmidrule(lr){9-11}
& \multirow{2}{*}{Entropy} & Spearman$_{0.5}$ & QA & SUM. & 0.801 & 0.795 & 0.713 & 0.634 & 0.637 & 4.580 \\
& & \cellcolor{gray!15}Random$_{3,2}$ & \cellcolor{gray!15}SUM. & \cellcolor{gray!15}QA & \cellcolor{gray!15}0.854 & \cellcolor{gray!15}0.843 & \cellcolor{gray!15}\textbf{0.705} & \cellcolor{gray!15}0.666 & \cellcolor{gray!15}0.634 & \cellcolor{gray!15}3.736 \\
\cmidrule(r){2-5} \cmidrule(lr){6-8} \cmidrule(lr){9-11}
& \multirow{2}{*}{JS-Div} & Spearman$_{1.0}$& QA & SUM. & 0.797 & 0.790 & 0.712 & \textbf{0.674} & \textbf{0.647} & \textbf{2.172} \\
& & \cellcolor{gray!15}--- & \cellcolor{gray!15}SUM. & \cellcolor{gray!15}QA & \cellcolor{gray!15}0.849 & \cellcolor{gray!15}0.835 & \cellcolor{gray!15}0.703 & \cellcolor{gray!15}0.702 & \cellcolor{gray!15}\textbf{0.636} & \cellcolor{gray!15}\textbf{2.029} \\
\cmidrule(r){2-5} \cmidrule(lr){6-8} \cmidrule(lr){9-11}
& \multirow{2}{*}{Sum} & Spearman$_{0.2}$ & QA & SUM. & 0.754 & 0.750 & \textbf{0.718} & 0.611 & 0.619 & 6.427 \\
& & \cellcolor{gray!15}Center$_{0.5}$ & \cellcolor{gray!15}SUM. & \cellcolor{gray!15}QA & \cellcolor{gray!15}0.800 & \cellcolor{gray!15}0.790 & \cellcolor{gray!15}0.695 & \cellcolor{gray!15}0.697 & \cellcolor{gray!15}0.632 & \cellcolor{gray!15}2.843 \\
\midrule
\multirow{8}{0.15\textwidth}{gemma-2-9b-it-bnb-4bit} & \multirow{2}{*}{CosSim} & --- & QA & SUM. & 0.841 & 0.830 & \textbf{0.793} & 0.608 & 0.591 & 6.299 \\
& & \cellcolor{gray!15}Center$_{0.2}$ & \cellcolor{gray!15}SUM. & \cellcolor{gray!15}QA & \cellcolor{gray!15}0.793 & \cellcolor{gray!15}0.778 & \cellcolor{gray!15}0.604 & \cellcolor{gray!15}0.758 & \cellcolor{gray!15}\textbf{0.637} & \cellcolor{gray!15}5.527 \\
\cmidrule(r){2-5} \cmidrule(lr){6-8} \cmidrule(lr){9-11}
& \multirow{2}{*}{Entropy} & Center$_{0.5}$ & QA & SUM. & 0.840 & 0.834 & 0.734 & \textbf{0.651} & \textbf{ 0.622} & \textbf{4.884} \\
& & \cellcolor{gray!15}Spearman$_{0.1}$ & \cellcolor{gray!15}SUM. & \cellcolor{gray!15}QA & \cellcolor{gray!15}0.749 & \cellcolor{gray!15}0.742 & \cellcolor{gray!15}0.624 & \cellcolor{gray!15}0.690 & \cellcolor{gray!15}0.612 & \cellcolor{gray!15}8.724 \\
\cmidrule(r){2-5} \cmidrule(lr){6-8} \cmidrule(lr){9-11}
& \multirow{2}{*}{JS-Div} & Center$_{0.2}$ & QA & SUM. & 0.772 & 0.764 & 0.756 & 0.591 & 0.593 & 8.566 \\
& & \cellcolor{gray!15}Spearman$_{0.2}$ & \cellcolor{gray!15}SUM. & \cellcolor{gray!15}QA & \cellcolor{gray!15}0.738 & \cellcolor{gray!15}0.732 & \cellcolor{gray!15}\textbf{0.626} & \cellcolor{gray!15}\textbf{0.777} & \cellcolor{gray!15}0.625 & \cellcolor{gray!15}\textbf{4.276} \\
\cmidrule(r){2-5} \cmidrule(lr){6-8} \cmidrule(lr){9-11}
& \multirow{2}{*}{Sum} & Center$_{0.1}$ & QA & SUM. & 0.732 & 0.725 & 0.783 & 0.597 & 0.606 & 6.531 \\
& & \cellcolor{gray!15}Spearman$_{\text{auto}}$ & \cellcolor{gray!15}SUM. & \cellcolor{gray!15}QA & \cellcolor{gray!15}0.745	 & \cellcolor{gray!15}0.736	 & \cellcolor{gray!15}0.617	 & \cellcolor{gray!15}0.749	 & \cellcolor{gray!15}0.625	 & \cellcolor{gray!15}5.901	 \\
\bottomrule \\
\end{tabular}
}
\caption{Best obtained results (w.r.t. Gap value) of AggTruth methods for hallucination detection, grouped by LLM, aggregation method, and task type.}
\label{tab:detection_results_all}
\end{table}

\begingroup

\setlength{\tabcolsep}{.03\textwidth}

\begin{table}
\centering
\scriptsize
\adjustbox{max width = \textwidth}{%
\begin{tabular}{lllccc}
\toprule
\rule{0pt}{2.5ex} \textbf{LLM} & \textbf{Source} & \textbf{Target} & \textbf{Test} & \textbf{Test (1)} & \textbf{Test (2)} \\
\midrule 
\multirow{2}{*}{llama-2-7b-chat-hf} & QA & SUM. & 0.752 & 0.670 & 0.739 \\
& \cellcolor{gray!15}SUM. & \cellcolor{gray!15}QA & \cellcolor{gray!15}0.722 & \cellcolor{gray!15}0.742 & \cellcolor{gray!15}0.684 \\
\midrule
\multirow{2}{*}{llama-3.1-8B-Instruct} & QA & SUM. & 0.886 & 0.689 & 0.722 \\
& \cellcolor{gray!15}SUM. & \cellcolor{gray!15}QA & \cellcolor{gray!15}0.715 & \cellcolor{gray!15}0.860 & \cellcolor{gray!15}0.857 \\
\midrule
\multirow{2}{*}{phi-3.5-mini-instruct} & QA & SUM. & 0.743 & 0.674 & 0.663 \\
& \cellcolor{gray!15}SUM. & \cellcolor{gray!15}QA & \cellcolor{gray!15}0.705 & \cellcolor{gray!15}0.730 & \cellcolor{gray!15}0.649 \\
\midrule
\multirow{2}{*}{gemma-2-9b-it-bnb-4bit} & QA & SUM. & 0.828 & 0.651 & 0.644 \\
& \cellcolor{gray!15}SUM. & \cellcolor{gray!15}QA & \cellcolor{gray!15}0.690 & \cellcolor{gray!15}0.790 & \cellcolor{gray!15}0.637 \\
\bottomrule \\
\end{tabular}
}
\caption{Best observed results in the conducted experiments for each LLM, task type, and test set.}
\label{tab:detection_best_results}
\end{table}

\endgroup

\end{document}